\documentclass[table]{bmvc2k}
\usepackage{bmvc2k_natbib}
\usepackage{amssymb} 

\title{PosBridge: Multi-View Positional Embedding Transplant for Identity-Aware Image Editing}

\addauthor{Peilin Xiong}{xiong-p@mm.inf.uec.ac.jp}{1}
\addauthor{Junwen Chen}{chen-j@mm.inf.uec.ac.jp}{1}
\addauthor{Honghui Yuan}{yuan-h@mm.inf.uec.ac.jp}{1}
\addauthor{Keiji Yanai}{yanai@cs.uec.ac.jp}{1}

\addinstitution{
Department of Informatics \\
 The University of Electro-Communications\\
 Tokyo, Japan
}

\runninghead{Xiong et al.}{PosBridge: Multi-View Positional Embedding Transplant}

\begin{document}

\maketitle

\begin{abstract}
Localized subject-driven image editing aims to seamlessly integrate user-specified objects into target scenes. As generative models continue to scale, training becomes increasingly costly in terms of memory and computation, highlighting the need for training-free and scalable editing frameworks.
To this end, we propose PosBridge—an efficient and flexible framework for inserting custom objects. A key component of our method is positional embedding transplant, which guides the diffusion model to faithfully replicate the structural characteristics of reference objects.
Meanwhile, we introduce the Corner Centered Layout, which concatenates reference images and the background image as input to the FLUX.1-Fill model. During progressive denoising, positional embedding transplant is applied to guide the noise distribution in the target region toward that of the reference object. In this way, Corner Centered Layout effectively directs the FLUX.1-Fill model to synthesize identity-consistent content at the desired location.
Extensive experiments demonstrate that PosBridge outperforms mainstream baselines in structural consistency, appearance fidelity, and computational efficiency, showcasing its practical value and potential for broad adoption.
\end{abstract}

\section{Introduction}

Subject-driven image generation aims to insert identity-consistent objects into target scenes using reference images or text prompts. It has broad applications in personalized content creation, virtual environments, and image editing. Existing approaches typically extract features from reference images and inject them into diffusion models via fine-tuning.

\begin{figure}[h]
\centering
\includegraphics[width=0.7\linewidth]{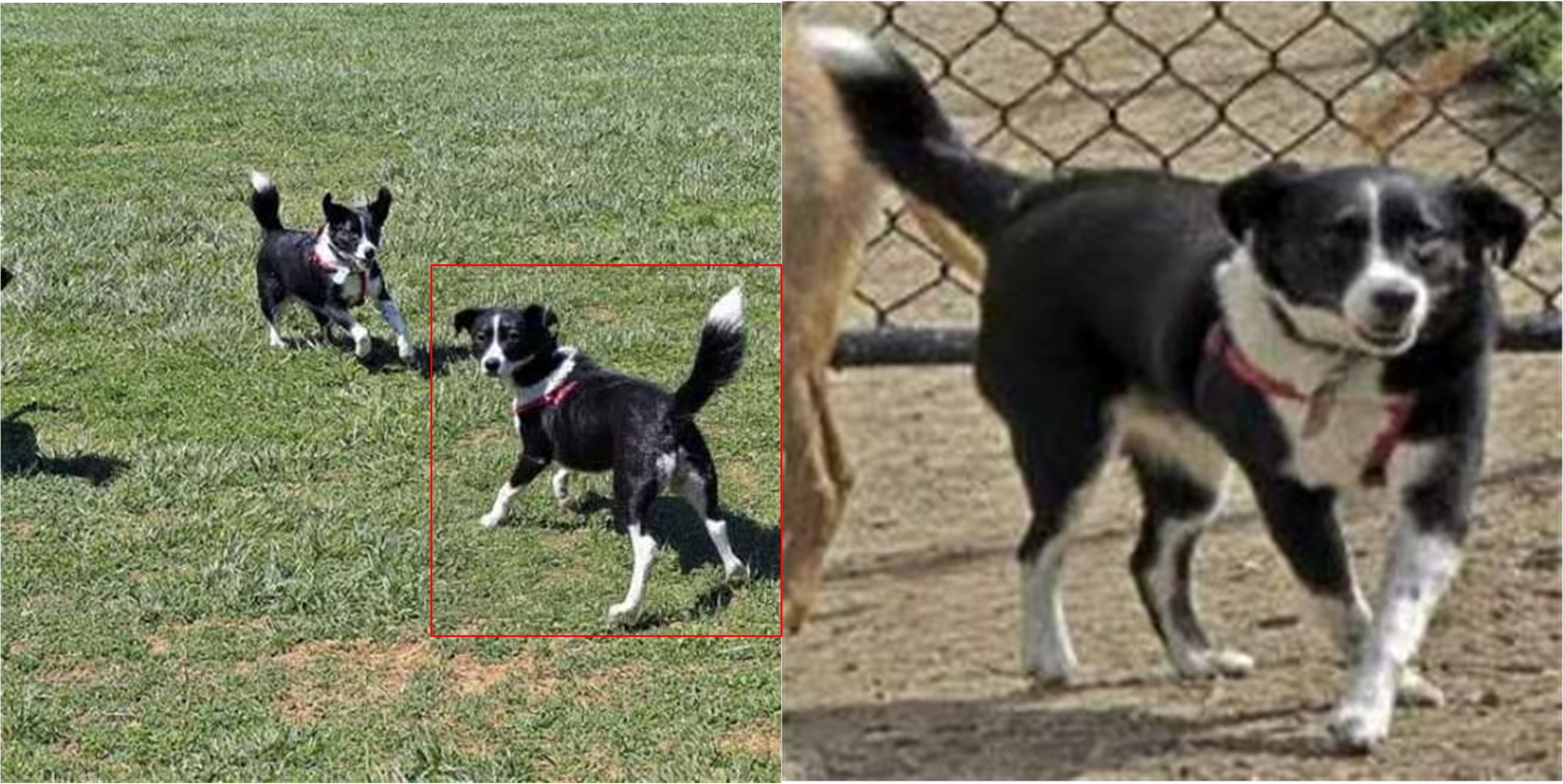}
\caption{A dog generated using our training-free method. Left: generated image. Right: reference image.}
\label{fig:example-doggrassssss}
\end{figure}

We propose \textbf{PosBridge}, a training-free framework for identity-aware and spatially controllable subject-driven image editing. Built on pre-trained MMDiT\cite{esser2024scaling} models, PosBridge achieves accurate object synthesis without requiring any dataset-specific training. Our method transplants positional embeddings from reference images to enforce structural alignment and adopts a Corner Centered Layout with token-level restructuring to mitigate reference bias and reduce computation. Figure~\ref{fig:example-doggrassssss} shows a representative result produced by our method, illustrating faithful identity preservation and seamless scene integration.

In particular, the core framework operates completely without training. An optional lightweight LoRA module can be incorporated to enhance fine-grained appearance fidelity. 
Extensive experiments demonstrate the effectiveness of our approach. We introduce PosBridge as a diffusion-based framework that enables structure-aware object insertion via positional embedding transplant, improves reference alignment through token restructuring in Corner Centered Layout, and optionally boosts visual realism using a class-agnostic LoRA enhancement module. Our method achieves state-of-the-art performance on both single- and multi-reference benchmarks in terms of semantic consistency and visual fidelity.

The main contributions are summarized as follows: 1) We propose \textbf{PosBridge}, a training-free editing framework that enables identity-consistent object insertion by transplanting positional embeddings from reference images. 2) We design a \textbf{Corner Centered Layout} with token-level restructuring to mitigate over-reliance on any single reference image, reducing copy-paste effects and improving object-scene integration. 3) We introduce a class-agnostic, lightweight \textbf{LoRA module} to boost detail realism in a parameter-efficient manner. 4) Our method achieves competitive or superior performance on semantic consistency and visual fidelity benchmarks, outperforming state-of-the-art baselines in both single- and multi-reference settings.

\section{Related Work}

\subsection{Positional Embedding in Vision Models}

Positional embeddings are critical for enabling spatial reasoning in transformer-based vision models. Early methods relied on absolute encodings, such as sinusoidal functions~\cite{vaswani2017attention}, which were not adaptive to variable input resolutions. Relative encodings improved generalization by modeling pairwise positional relationships. More recently, techniques such as Rotary Positional Embedding (RoPE)~\cite{su2024roformer} encode position as rotations in attention space, offering enhanced spatial modeling capabilities.

\subsection{Subject-Driven Image Editing}

Subject-driven image editing involves inserting identity-preserving objects into new scenes. Early approaches like DreamEdit~\cite{li2023dreamedit} and Cones 2~\cite{liu2023customizable} rely on iterative diffusion or instance-specific token embeddings for personalized synthesis. More recent zero-shot methods, such as AnyDoor~\cite{chen2024anydoor}, IMPRINT~\cite{song2024imprint}, and BIFRÖST~\cite{li2024bifrost}, introduce mask-free or 3D-aware pipelines that improve compositional realism and spatial flexibility. Despite progress, achieving structural control, semantic alignment, and natural blending simultaneously remains a challenge. Our method addresses these limitations by leveraging positional embedding transplant and the Corner Centered Layout to enable structure-aware and editable object integration.

\subsection{Modifying RoPE in Diffusion Models}

Two recent training-free approaches manipulate diffusion transformers’ positional representations to control generation. \textit{RoPECraft} warps the rotary positional embeddings (RoPE) of a diffusion transformer using dense optical flow from a reference video and then aligns motion trajectories during denoising~\cite{gokmen2025ropecraft}. It also employs a Fourier-based regularization to suppress high-frequency artifacts and achieve realistic motion transfer without fine-tuning~\cite{gokmen2025ropecraft}. \textit{GrounDiT}, targeting spatial grounding, modifies RoPE in a different way: it enforces \emph{semantic sharing} by assigning the same RoPE to tokens belonging to a bounding-box region, causing separate patches and the full image to share positional embeddings and thus become semantic clones~\cite{lee2024groundit}. These aligned RoPEs enable the model to generate and denoise patches jointly and transplant them back into the main image at each denoising step~\cite{lee2024groundit}, providing precise spatial control. Together, these works show that careful manipulation of rotary positional embeddings, whether warping them with optical flow or sharing them among region tokens, can endow diffusion models with motion or layout control without retraining.

\begin{figure}[h]
\centering
\includegraphics[width=0.95\linewidth]{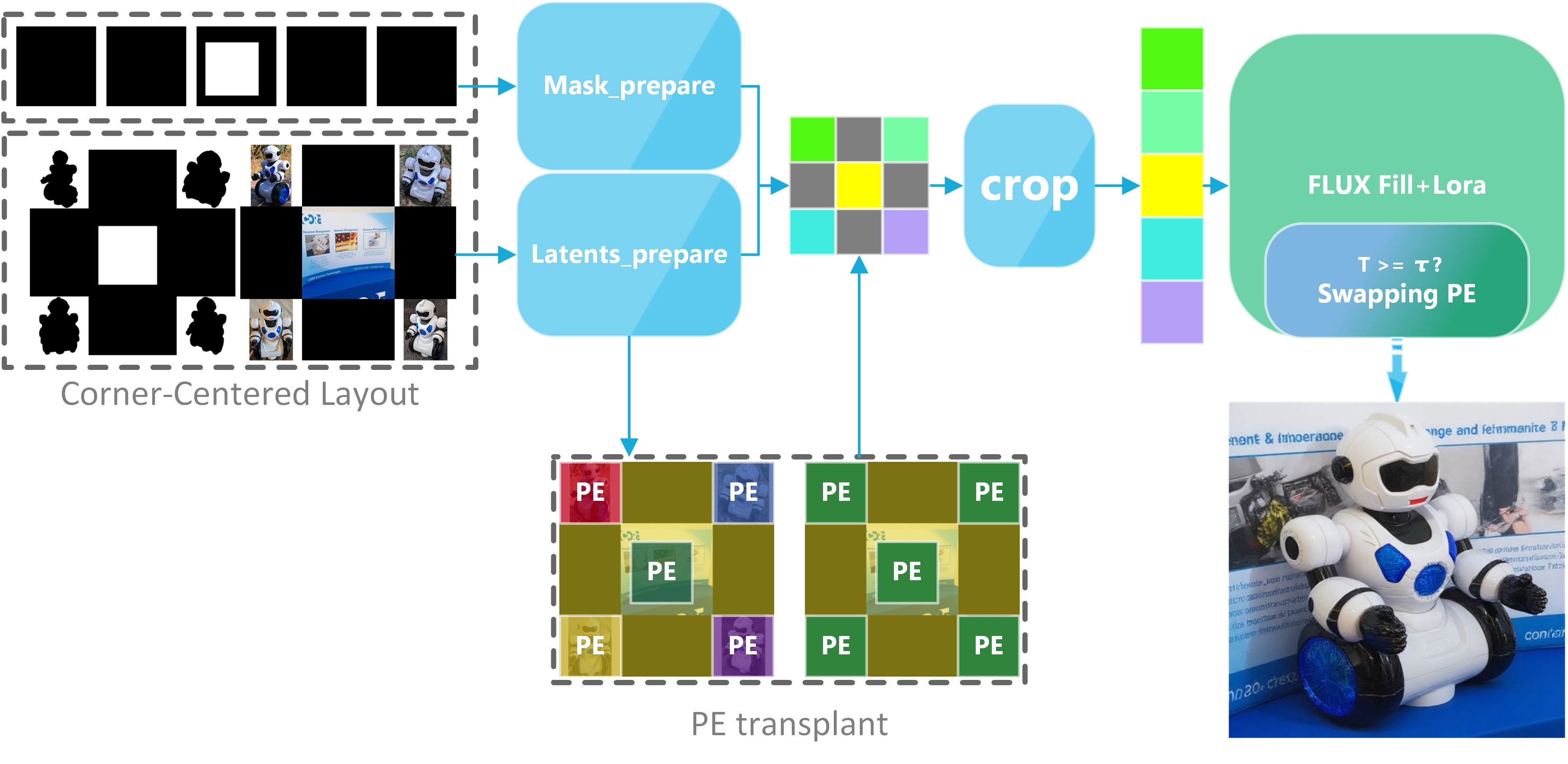}
\caption{Overview of our PosBridge pipeline.}
\label{fig:pipeline-edit}
\end{figure}

\section{Method}
\begin{figure}[h]
\centering
\includegraphics[width=0.9\linewidth]{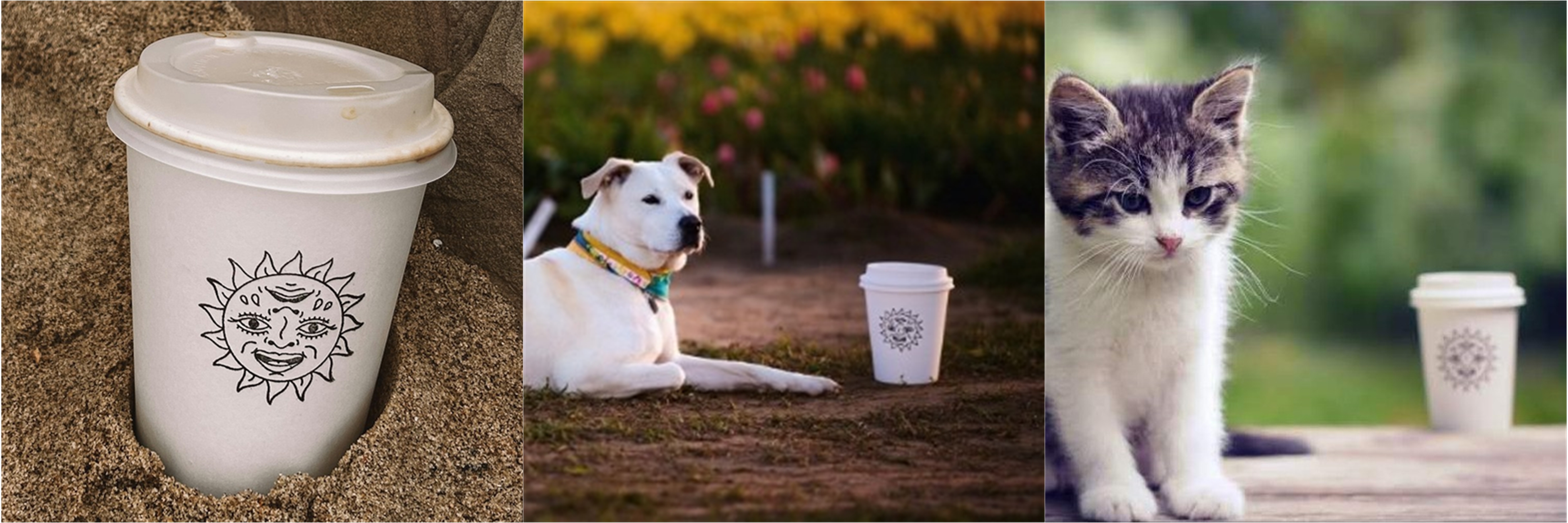}
\caption{Result demonstrating seamless integration of the inserted object with consistent illumination and geometry.}
\label{fig:example-cupdog}
\end{figure}

FLUX.1-Fill~\cite{labs2025flux} enables localized image editing by using a binary mask to restrict modifications while preserving the unedited background. Building on this foundation, we propose a training-free, subject-driven editing framework that transplants positional embeddings and leverages reference images arranged in Corner Centered Layout to guide the synthesis of identity-consistent objects at specified locations.
Figure \ref{fig:pipeline-edit} summarizes the overall workflow, and Figure \ref{fig:example-cupdog} illustrates a typical result. The generated composition faithfully preserves the cup’s identity while seamlessly matching the scene’s depth, focus blur, and angled sunset illumination, demonstrating the framework’s ability to reconcile object appearance with its visual context.

\subsection{Mask-Guided Positional Embedding Transplant}

\begin{figure}[h]
    \centering
    \begin{minipage}[t]{0.25\linewidth}
        \centering
        \includegraphics[width=\linewidth]{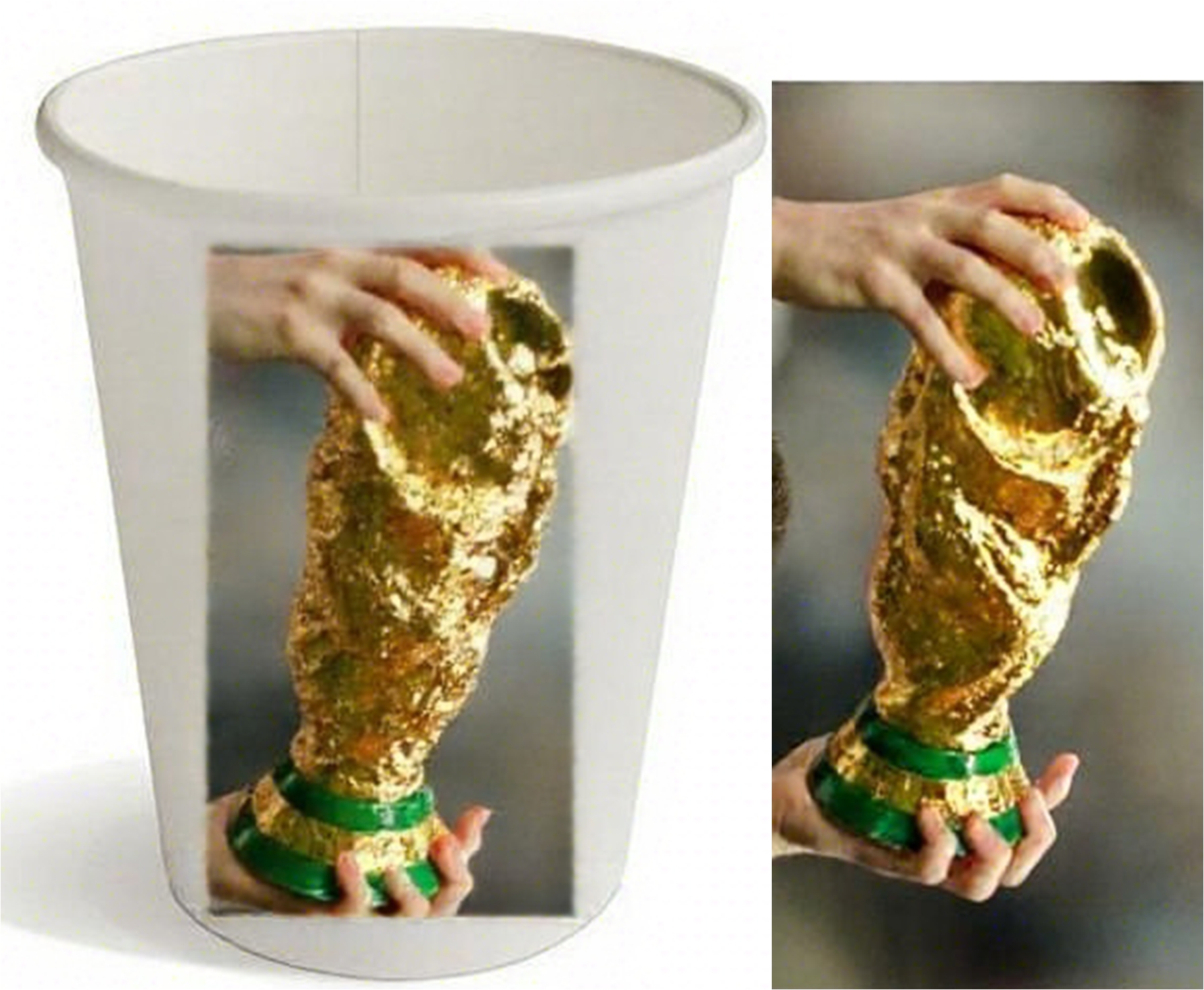}
        \label{fig:img_swap}
        \small (a)
        
    \end{minipage}
    \hfill
    \begin{minipage}[t]{0.25\linewidth}
        \centering
        \includegraphics[width=\linewidth]{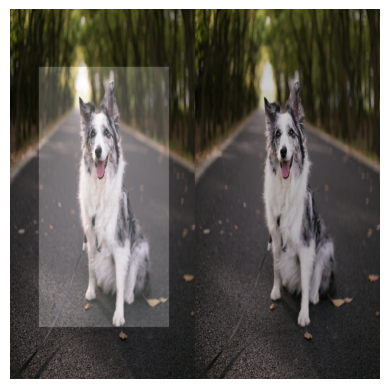}
        \label{fig:same-dog}
        \small (b)
        
    \end{minipage}
    \hfill
    \begin{minipage}[t]{0.25\linewidth}
        \centering
        \includegraphics[width=\linewidth]{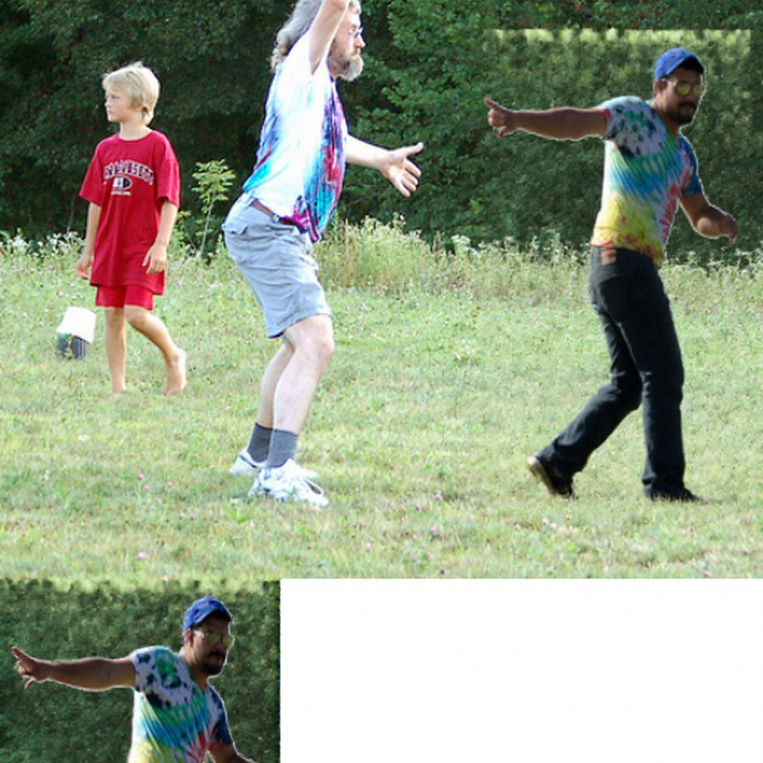}
        \label{fig:bord}
        \small (c)
        
    \end{minipage}
    \caption{(a) Shared positional embeddings produce duplicated content across spatial locations. (b) Spatial priors dominate over textual prompts. (c) Spatial continuity across boundaries influences diffusion output.}
    \label{fig:pos_embedding_effects}
\end{figure}

\paragraph{Positional Embeddings Govern Structural Transfer.}
When different spatial regions are assigned identical positional embeddings, the diffusion model tends to generate nearly identical content in both locations. This behavior suggests that structural information is primarily encoded by positional embeddings, rather than by absolute spatial coordinates. As illustrated in Figure~\ref{fig:pos_embedding_effects}(a), duplicating the positional embeddings from region A to region B leads the model to synthesize highly similar structures in both regions, effectively mirroring the content.

\paragraph{Spatial Priors Override Textual Semantics.}
Figure~\ref{fig:pos_embedding_effects}(b) illustrates that when two identical images are concatenated and one half is partially masked, FLUX.1-Fill often reconstructs the masked region by copying from the unmasked half, even when the textual prompt suggests a different object. For example, a prompt indicating a \textit{cat} may still yield reconstruction of a \textit{dog} from the reference image, revealing that visual priors can dominate over prompt conditioning.

\paragraph{Swapping Positional Embeddings.}  
Building on these insights, we \textit{transplant the positional embeddings (PE) of the target (editable) region onto the reference (non-editable) region} during the early denoising steps ($t \ge \tau$). This swap pushes the target’s diffusion trajectory to follow that of the reference, enforcing global structural alignment. After $\tau$ steps we restore the original PE, releasing the target from this constraint. By then the target region already mirrors the coarse geometry of the reference object, allowing the model to refine appearance details using reference features while blending them naturally with the surrounding context.

This \textit{temporally-gated transplantation} offers complementary control:  
(i) an early phase that provides strong global guidance through PE swapping, and  
(ii) a late phase that, under the native PE, performs local context refinement.  
Together they yield coherent, editable outputs in which newly generated content is seamlessly woven into the original scene.

\paragraph{Detail Enhancement via LoRA (Optional).}
\label{sec:method}
To further enhance visual fidelity, we optionally incorporate a lightweight LoRA module~\cite{hu2022lora}, inspired by DreamBooth\cite{ruiz2023dreambooth}. Unlike conventional instance-specific fine-tuning, we train a shared, class-agnostic LoRA module using distinct textual tokens to represent different categories. This class-conditioned design enables FLUX.1-Fill to synthesize fine-grained object details while maintaining broad generalization. Importantly, our positional embedding transplant mechanism remains fully compatible with the LoRA-enhanced model, highlighting its modularity and potential for future extensions.

\paragraph{Impact of $\tau$ on Structure–Context Trade-off.}
The threshold parameter $\tau$ is critical for balancing structural control and scene coherence. If $\tau$ is set too low—allowing transplanted embeddings to persist too long—artifacts such as ghosting or poor blending may emerge. This issue becomes more pronounced in multi-reference scenarios, where conflicting spatial priors can destabilize generation. Therefore, $\tau$ must be carefully calibrated to ensure effective structural transfer during early denoising, while enabling smooth contextual integration in later stages.

\subsection{Corner Centered Layout and Token Restructuring}

\paragraph{FLUX.1-Fill Input Representation.}
FLUX.1-Fill requires a background image $I$ and a binary edit mask $M \in \{0,1\}^{H \times W}$ that specifies the editable region (with white pixels indicating insertion areas). Based on these inputs, the model receives four input streams: a \texttt{noise\_token}, initialized with Gaussian noise; an \texttt{image\_token}, obtained by VAE-encoding the masked background $I \odot M$; a \texttt{mask\_token}, derived by flattening the binary mask with a stride of $16 \times 16$; and a \texttt{text\_token}, generated from T5-encoded prompts and appended as a global conditioning token.

The first three streams are reshaped into spatial grids of size $H/16 \times W/16$, with channel dimensions of 64 for both \texttt{noise\_token} and \texttt{image\_token}, and 256 for \texttt{mask\_token}. These spatial tokens are concatenated along the channel axis to form a 384-channel tensor, which is then linearly projected into a 3072-dimensional feature representation. This projected tensor, together with the non-spatial \texttt{text\_token}, is passed to the FLUX.1-Fill backbone for conditioned generation.

\paragraph{Corner Centered Layout.}
To mitigate the spatial bias in self-attention, where attention weights decay with distance, we place four reference images at the corners of the canvas and position the background image in the center. This arrangement ensures that the editable region remains approximately equidistant from all reference patches, reducing the model’s tendency to over-attend to nearby sources. Each reference image is first cropped using its segmentation mask to isolate the target object, then resized and padded to match the size and location of the designated corner slot. Corresponding masks are constructed using the inverse segmentation mask to suppress background influence during encoding of the \texttt{image\_token}. Empirically, we find that generating an all-black mask (disabling edit regions) for the reference \texttt{mask\_token} leads to improved integration between the inserted object and the surrounding scene.
This design also alleviates the issue illustrated in Fig.~\ref{fig:pos_embedding_effects}(c), where the spatial continuity across region boundaries can undesirably influence the diffusion output. By anchoring the references to the corners and isolating them from the editable region, we reduce boundary leakage and improve generation robustness.

\paragraph{Token Cropping and Restructuring.}
To reduce memory usage and computational cost, we reshape the \texttt{image\_token} into a $H/16 \times W/16$ grid, crop five informative regions (four corners and the center), and concatenate them into a compact latent sequence. The same cropping and restructuring procedure is applied to the positional embeddings to maintain spatial correspondence. Although the resulting token sequence differs from the original canvas arrangement, it is functionally equivalent to a arranging regions diagonally (from top-left to bottom-right). Since FLUX.1-Fill’s generation is primarily guided by positional embeddings rather than the literal order of tokens, this simplification has a negligible impact on visual quality while significantly improving memory efficiency.

\section{Experimental Setup}
\label{sec:experiment}

\paragraph{Baselines.}  
We evaluate PosBridge against two representative methods. AnyDoor~\cite{chen2024anydoor} injects identity and appearance features into a diffusion model for high-fidelity object insertion. Insert-Anything~\cite{song2026insertanything}, which shares the same FLUX.1-Fill backbone as ours, integrates FLUX.1-Redux’s image encoder with FLUX.1-Fill’s generator to complete the input of reference information. Notably, both Insert-Anything and our method utilize the same FLUX.1-Fill backbone with identical masks to remove background from the reference image.

\paragraph{Metrics.}  
We use two cosine similarity-based metrics for semantic alignment: \textbf{CLIP-I Score}~\cite{radford2021clip} (CLIP image encoder) and \textbf{DINOv2 Score}~\cite{oquab2023dinov2} (self-supervised ViT features). Their average forms the \textbf{Composite Score (C+D)}. Evaluations are conducted on the DreamBooth dataset~\cite{ruiz2023dreambooth}, using class-wise foreground masks to extract features from both reference and generated regions. Cosine similarity is computed between each generated region and its class prototype, with final scores averaged across all categories.

\paragraph{Implementation.}  
All experiments are performed with FLUX.1-Fill on a single NVIDIA A100 GPU at $512 \times 512$ resolution. LoRA is trained for 6500 steps (rank 32) using rare T5 tokens (e.g., \texttt{"Kwa"}, \texttt{"McKe"}) as class prompts. During inference, we set the positional embedding swap threshold to $\tau = 2$.

\subsection{Quantitative Results}

The quantitative results in Table~\ref{tab:quant_results_combined} show that \textbf{Ours (LoRA + SwPE)} achieves the best overall trade-off between semantic alignment and reference fidelity, with the highest \textbf{CLIP-I} score (\textbf{0.8827}), indicating strong instance-level consistency with the reference image. Although \textbf{Insert-Anything} attains the highest DINOv2 score (\textbf{0.7638}), this may indicate over-reliance on reference structure, as suggested by visible copy-paste effects in the qualitative analysis. While it preserves structure well, it often shows limitations in natural blending. In contrast, \textbf{AnyDoor} performs well in general zero-shot insertion, but its lower CLIP-I and composite scores point to weaker correspondence with reference identity.

\begin{table}[h]
\centering
\caption{Quantitative comparison on semantic fidelity metrics. Results are averaged over four random seeds (42, 100, 200, 600). Methods marked with * use single-reference input. Ablation variants isolate the contributions of LoRA and Swapping Positional Embeddings (SwPE).}
\label{tab:quant_results_combined}
\resizebox{0.8\linewidth}{!}{%
\begin{tabular}{|l|c|c|c|c|}
\hline
\textbf{Method} & \textbf{DINOv2 $\uparrow$} & \textbf{CLIP-I $\uparrow$} & \textbf{C+D $\uparrow$} & \textbf{Training-Free} \\
\hline
AnyDoor*~\cite{chen2024anydoor}         & 0.7247            & 0.8662            & 0.7954          &  \\
Insert-Anything*~\cite{song2026insertanything} & \textbf{0.7638}   & 0.8777            & \textbf{0.8208} &  \\
\rowcolor[HTML]{EAEAEA}
Copy-Paste                  & 0.7760            & 0.8882            & 0.8321          &  \\
\hline
Ours (SwPE)               & 0.7002            & 0.8701            & 0.7852          & \checkmark  \\
Ours (LoRA + SwPE)*            & 0.7149            & 0.8778           & 0.7963          &  \\
Ours (LoRA)             & 0.7264            & 0.8815            & 0.8039          &  \\
Ours (LoRA + SwPE)             & 0.7351            & \textbf{0.8827}   & 0.8089          &  \\

\hline
\end{tabular}
}
\end{table}

\begin{figure}[h]
\centering
\includegraphics[width=0.9\linewidth]{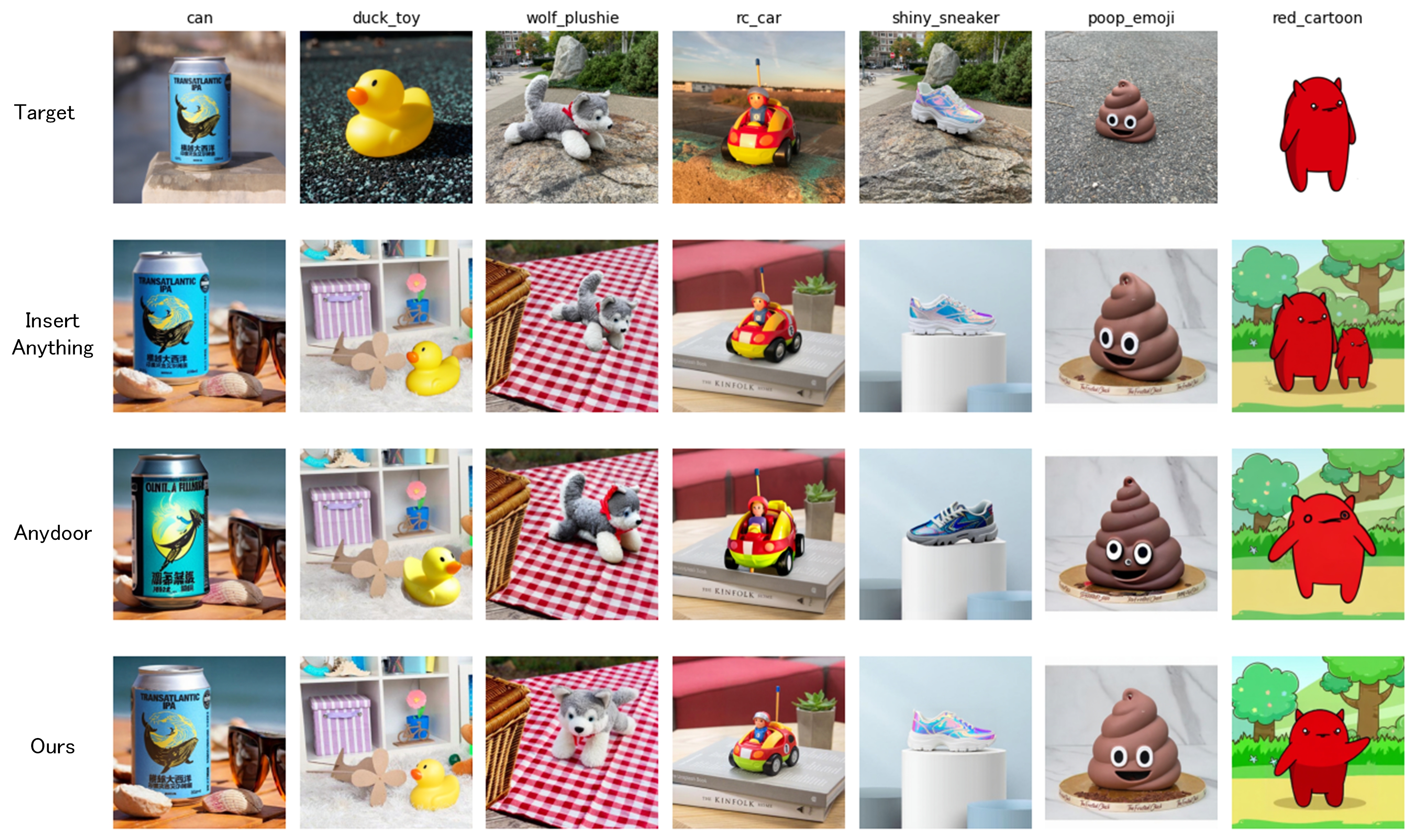}
\caption{Qualitative comparison across methods. Our approach yields better shape, shading, and scene consistency.}
\label{fig:qualitative}
\end{figure}
\begin{figure}[h]

\centering
\includegraphics[width=0.9\linewidth]{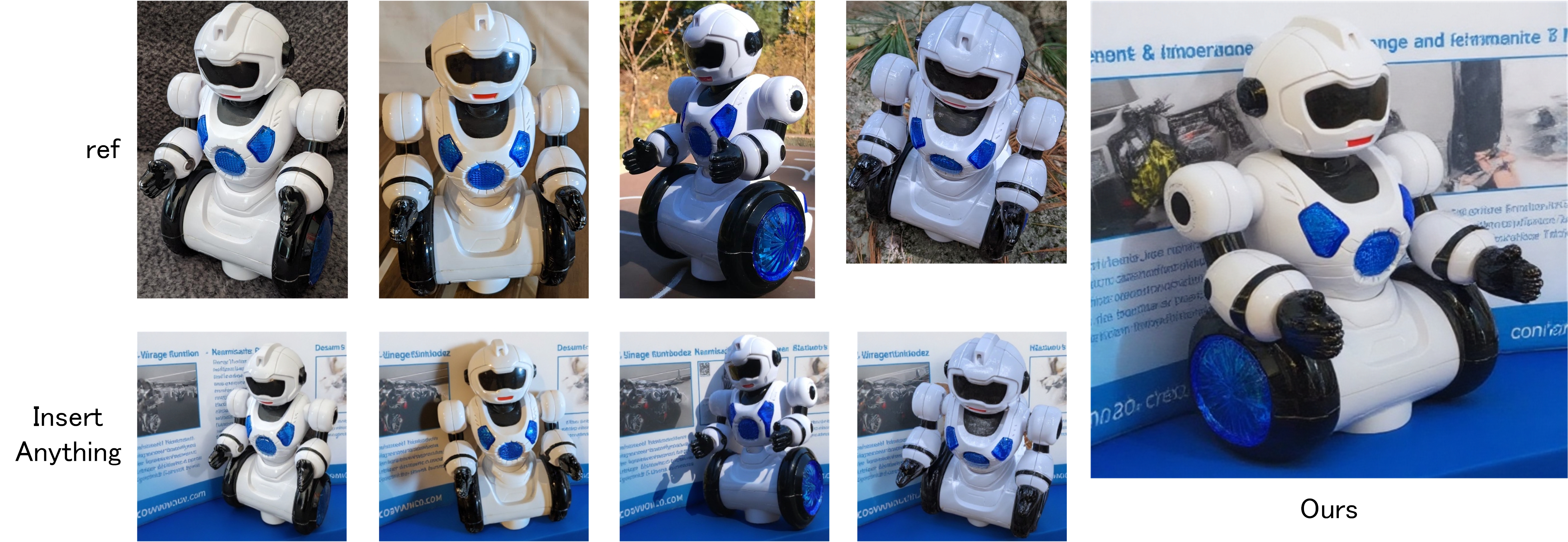}
\caption{Insert-Anything shows copy-paste effects (e.g., lighting, pose), while ours achieves better scene-aware integration.}
\label{fig:IAOurs}
\end{figure}

\begin{figure}[h]
\centering
\includegraphics[width=0.9\linewidth]{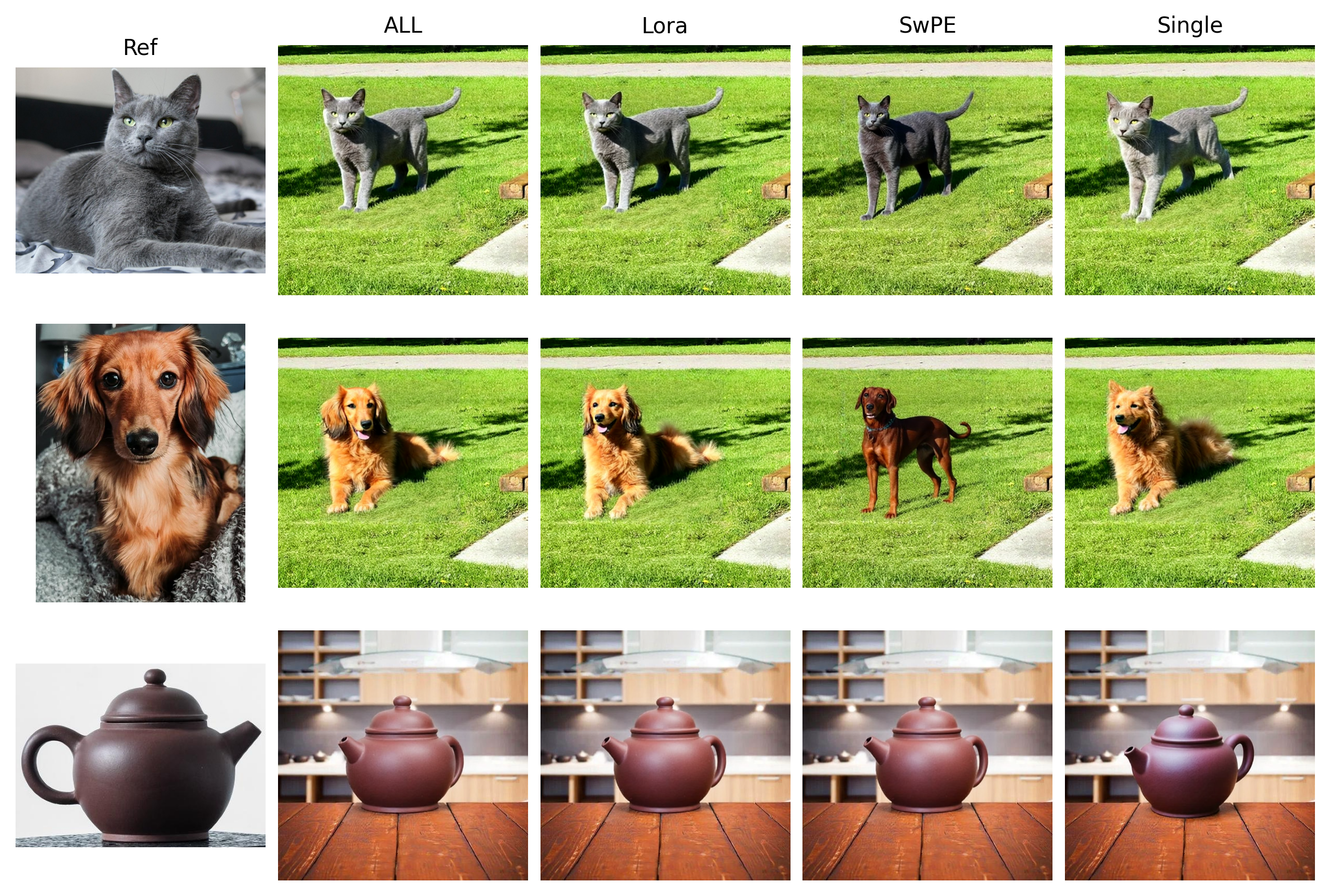}
\caption{Ablation Study on LoRA and SwPE. We evaluate three variants to assess component contributions: (1) LoRA + SwPE (full model, denoted as “ALL”), (2) LoRA Only, and (3) SwPE Only. As visualized, SwPE and Single (i.e., baseline) exhibit clear differences from the full model. These results demonstrate the complementary effects of LoRA and SwPE.}
\label{fig:Ablation}
\end{figure}
\subsection{Qualitative Analysis}

We provide qualitative comparisons to highlight the benefits of our method in terms of consistency, realism, and controllability.
In single-reference settings, generated objects frequently fail to harmonize with the background scene—exhibiting shape distortion, inconsistent shading, or identity collapse under novel poses and lighting conditions. \textbf{Insert-Anything}, in particular, suffers from visible copy-paste effects: objects are directly replicated from the reference, often retaining source-specific lighting and texture without adapting to the target background, as shown in Figure~\ref{fig:IAOurs}. For example, the robot’s surface reflections in each Insert-Anything result closely match those in the reference image; in the second column, the yellow light source is even directly copied without any adaptation to the new scene. The robot’s orientation is also unchanged, further indicating a lack of contextual blending.
In contrast, our method leverages four reference views and spatially guided synthesis via positional embedding transplant. This design enables the model to form a more robust and pose-consistent object representation. Consequently, the generated outputs demonstrate improved geometric coherence, viewpoint consistency, and seamless integration with the surrounding scene.

\begin{figure}[h]
\centering
\includegraphics[width=0.9\linewidth]{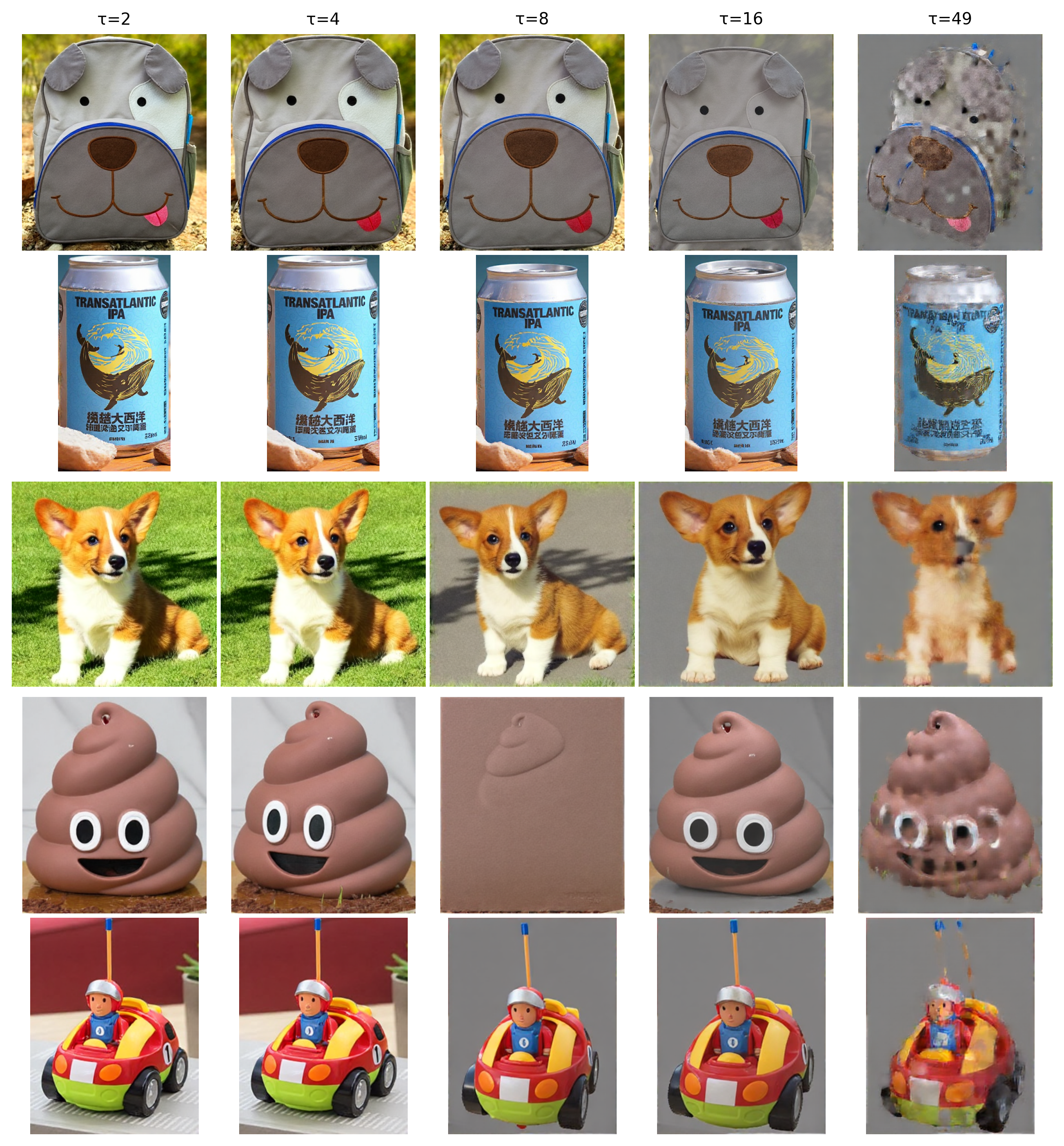}
\caption{Ablation study on the temporal threshold $\tau$. Columns (left $\rightarrow$ right) correspond to $\tau=2,4,8,16,49$. Small $\tau$ values yield clean, well-blended results, whereas larger $\tau$ values introduce desaturated backgrounds and reference ghosting.}
\label{fig:Ablation_t}
\end{figure}

\subsection{Ablation Studies}

We evaluate three variants to analyze the contribution of each component:  
(1) \textbf{LoRA + SwPE} (full model),  
(2) \textbf{LoRA}, and  
(3) \textbf{SwPE}.  

As shown in Table~\ref{tab:quant_results_combined}, LoRA improves fine-grained detail and identity consistency (higher CLIP-I), while SwPE enhances spatial structure (higher DINOv2). Combining both yields the best overall performance, confirming their complementary roles.

Figure~\ref{fig:Ablation} provides qualitative comparisons. The \textbf{ALL} configuration consistently produces the most realistic and coherent results. The \textbf{LoRA} variant lacks geometric precision (e.g., in eyes, ears, or spout details), while the \textbf{SwPE} variant preserves layout better but shows degraded identity. The \textbf{Single} variant (using one reference without Corner Centered Layout) often exhibits copy-paste effects and shows reduced generalization capability.

These findings indicate that LoRA and SwPE complement each other: LoRA enhances semantic fidelity, while SwPE improves spatial structure. Their combination, along with multi-reference input, leads to robust and high-quality image generation.

\paragraph{Effect of $\tau$ on Structural Guidance.}  
The temporal threshold $\tau$ controls the duration of positional embedding transplant during early denoising. Figure~\ref{fig:Ablation_t} illustrates its impact:

\begin{itemize}
  \item \textbf{$\tau=2$, $4$}: Optimal trade-off, producing sharp images with faithful geometry and semantic details. At $\tau=4$, minor background leakage may occur (e.g., residual grass around objects).
  
  \item \textbf{$\tau=8$}: Degraded blending with visible boundary artifacts and inconsistent scene integration.
  
  \item \textbf{$\tau=16$, $49$}: Severe quality loss including desaturated backgrounds, poor fusion, and reference ghosting, indicating failed structural control.
\end{itemize}

These observations confirm that \textit{brief structural guidance} ($\tau \leq 4$) best balances alignment and blending. We use $\tau=2$ in all experiments for reliable identity transfer with minimal artifacts.

\section{Conclusion and Future Work}
We present \textbf{PosBridge}, a training-free framework for high-fidelity subject insertion that leverages positional embedding transplant and Corner Centered Layout. Without requiring any additional training, PosBridge enables accurate and controllable object synthesis. Extensive experiments demonstrate that PosBridge outperforms existing methods in both structural consistency and generation robustness. When combined with a lightweight shared LoRA module, the two components complement each other, enhancing structural alignment and appearance fidelity, respectively.

\bibliography{references}
\end{document}